\documentstyle[aaai]{article}

\newtheorem{ex}{Example}

\newtheorem{definition}{Definition}
\newtheorem{theorem}{Theorem}

\newlength{\axiomwidth}
\newcommand{\PR}{\mbox{$\ \vdash\ $}}

\newcommand{\E}{{\cal E}}
\newcommand{\TurnOn}{{\mbox{\textit{TurnOn}}}}

\newcommand{\Running}{{\mbox{\textit{Running}}}}
\newcommand{\Petrol}{{\mbox{\textit{Petrol}}}}
\newcommand{\Fill}{{\mbox{\textit{Fill}}}}
\newcommand{\InjectA}{{\mbox{\textit{InjectA}}}}
\newcommand{\InjectB}{{\mbox{\textit{InjectB}}}}
\newcommand{\InjectC}{{\mbox{\textit{InjectC}}}}
\newcommand{\InjectD}{{\mbox{\textit{InjectD}}}}
\newcommand{\InjectE}{{\mbox{\textit{InjectE}}}}
\newcommand{\Infected}{{\mbox{\textit{Infected}}}}
\newcommand{\Protected}{{\mbox{\textit{Protected}}}}
\newcommand{\TypeO}{{\mbox{\textit{TypeO}}}}
\newcommand{\TypeA}{{\mbox{\textit{TypeA}}}}
\newcommand{\Weak}{{\mbox{\textit{Weak}}}}
\newcommand{\Strong}{{\mbox{\textit{Strong}}}}
\newcommand{\Bite}{{\mbox{\textit{Bite}}}}
\newcommand{\Expose}{{\mbox{\textit{Expose}}}}

\newcommand{\Initiation}{{\mbox{\textit{Initiation}}}}
\newcommand{\HoldsAt}{{\mbox{\textit{HoldsAt}}}}
\newcommand{\Termination}{{\mbox{\textit{Termination}}}}
\newcommand{\HappensAt}{{\mbox{\textit{HappensAt}}}}

\newcommand{\PGe}{{\mbox{\textit{PG}}}}
\newcommand{\NGe}{{\mbox{\textit{NG}}}}
\newcommand{\PA}{{\mbox{\textit{PA}}}}
\newcommand{\NA}{{\mbox{\textit{NA}}}}
\newcommand{\PP}{{\mbox{\textit{PP}}}}
\newcommand{\NP}{{\mbox{\textit{NP}}}}

\newcommand{\ARE}{{\mathcal{A}_{\E}}}
\newcommand{\PED}{{P_{\E}(D)}}
\newcommand{\NPED}{{P_{\E}(D'\cup \Delta_w)}}
\newcommand{\NNPED}{{P_{\E}(D''\cup \Delta)}}

\newcommand{\Empty}{Empty}

\begin{document}

\title{Planning with Incomplete Information}

\author{Antonis Kakas \\
University of Cyprus \\
{\tt antonis@cs.ucy.ac.cy} 
\And
Rob Miller \\
University College, U.K. \\
{\tt rsm@ucl.ac.uk} 
\And
Francesca Toni\\
Imperial College, U.K. \\
{\tt ft@doc.ic.ac.uk} 
}

\maketitle

\begin{abstract}
\begin{quote}
Planning is a natural domain of application for frameworks of
reasoning about actions and change. In this paper we study how
one such framework, the Language ${\cal E}$, can form the basis
for planning  under (possibly) incomplete information.
We define two types of plans: {\em weak} and {\em safe} plans, and
propose a planner, called the {\em ${\cal E}$-Planner}, which is 
often able to extend an initial weak plan into a safe plan even 
though the (explicit) information available is incomplete,
e.g. for cases where the initial state is not completely known.
The ${\cal E}$-Planner is based upon a reformulation of the
Language ${\cal E}$ in argumentation terms and a natural proof theory
resulting from the reformulation.
It uses an extension of this proof 
theory by means of abduction for the generation of plans and 
adopts argumentation-based techniques for extending weak plans
into safe plans. We provide representative examples illustrating 
the behaviour of the ${\cal E}$-Planner,
in particular for cases where the status of fluents is incompletely known.
\end{quote}
\end{abstract}

\section{Introduction}

General formalisms of action and change can provide a natural framework
for the problem of planning. They can offer a high level of 
expressivity and a basis for the development of general 
purpose planning algorithms. 

We study how one such formalism, 
the Language ${\cal E}$ \cite{KaMi97,KaMi97bis}, 
can form a basis for planning. 
To do this we exploit the reformulation \cite{lpnmr99} of the 
Language ${\cal E}$ into an argumentation framework 
and the associated proof theory offered by this 
reformulation. A simple extension of this
argumentation-based  proof theory with abduction 
forms the basis of planning algorithms 
within the framework of the  Language ${\cal E}$.

In this paper we will be particularly interested in 
addressing the specific problem of planning 
under incomplete information. 
This amounts to  planning in cases where some 
information is missing,  as for example when we do not have full knowledge of the 
initial state of the problem. In general, we assume that this missing 
information cannot be ``filled in'' by additional actions 
in the plan as it may refer to properties that cannot be affected
by any type of action in the theory or to an initial time 
before which no actions can be performed. 
Instead, the planner needs to be able to reason despite
this incompleteness and construct plans 
where this luck of information 
does not matter for achieving the final goal.

We define a planner, call the ${\cal E}$-Planner, 
which is able to solve this type of planning problems with
incomplete information. It works by first generating a conditional 
plan based on one possible set of arguments in the corresponding
argumentation theory of the planning domain. These plans 
are called weak plans and may not be successful under 
every possibility for the missing information. 
The planner then uses 
further argumentation reasoning to extend the weak plan 
to a safe plan which is able to achieve the planning goal 
irrespective of the particular status of the missing imformation.

Planning under incomplete information is a relatively
new topic.  In \cite{reiter99} this problem is called 
``Open World Planning'' and is studied within the framework of the
situation calculus. The incomplete information refers 
to the initial situation of the problem and a theorem 
prover is used to reason about properties at this situation.
Other related work on planning within formal frameworks for 
reasoning about actions and change is \cite{levesque96}, which
defines a notion of conditional plans,  
\cite{shanahan97,denecker92}, with a formulation of abductive planning in the
event calculus and \cite{dimopoulos97,lifchitz99}, which 
study ``answer set planning'' within 
extended logic programming. 

\section{A Review of the Basic Language ${\cal E}$}\label{RevSection}
\label{E-language}

The Language ${\cal E}$ is really a 
collection of languages. The particular vocabulary of each language 
depends on the domain being represented, but always includes a set of 
{\it fluent constants}, a set of {\it  action constants}, and a partially
ordered set 
$\langle \Pi , \preceq \rangle$ of {\it time-points}. 
For this paper where we are interested in linear planning we 
will assume that $\preceq$ is a total order.
A {\it fluent literal}
is either a fluent constant $F$ 
or its negation $\neg F$. 
%

{\em Domain descriptions} in the Language ${\cal E}$ are collections
of  statements of three kinds
(where $A$ is an action constant, $T$ is a time-point, 
$F$ is a fluent constant, $L$ is a fluent literal
and  $C$ is a set of
fluent literals):
{\em t-propositions} (``t'' for ``time-point''),
of the form
$L \mbox{ {\tt holds-at} } T$;
{\em h-propositions} (``h'' for ``happens''),  
of the form 
$A \mbox{ {\tt happens-at} } T$;
{\em c-propositions} (``c'' for ``causes''), of the form
$A \mbox{ {\tt initiates} } F  \mbox{ {\tt when} } C$  or  
$A \mbox{ {\tt terminates} } F  \mbox{ {\tt when} } C$.
%
When $C$ is empty, the c-propositions are written as 
``$A \mbox{ {\tt initiates} } F$'' and 
``$A \mbox{ {\tt terminates} } F$'', resp. 

The semantics of ${\cal E}$ is based on 
simple definitions of interpretations,
defining the truth value of t-propositions at
each particular time-point,
and models.
Briefly, (see \cite{KaMi97bis,KaMi97} for more details) these are given as 
follows:

\begin{itemize}

\item An {\em interpretation} is a mapping $H: \Phi \times \Pi \mapsto \{true,false\}$,
where $\Phi$ is the  set of fluent constants and $\Pi$ is  the set of time-points in ${\cal E}$.
Given a set of fluent literals $C$ and a 
time-point $T$, an interpretation $H$ {\em satisfies} 
$C$ {\em at} $T$ iff for each fluent constant $F \in C$, 
$H(F,T)=true$, and for each fluent constant 
$F'$ such that $\neg F' \in C$, $H(F',T)=false$.

\item 
Given a time-point $T$, a fluent constant $F$ and an interpretation $H$,
$T$ is an {\em  initiation-point} ({\em ter\-mi\-na\-tion-point} resp.)
{\em for} $F$ {\em in} $H$ {\em relative to} a domain description $D$ 
iff there is an action constant
$A$ such that (i) $D$ contains both an h-proposition $A$ {\tt happens-at} $T$ 
and a
c-proposition 
$A$ {\tt initiates} ({\tt terminates}, resp.) $F$ {\tt when} $C$,
and (ii) $H$ satisfies $C$ at $T$.
Then, an interpretation $H$ is a {\em model} of a given domain description $D$
iff, for every fluent constant $F$ and time-points $T_{1}\prec T_{3}$: 

\begin{enumerate}
\item If there is no initiation- or termination-point
$T_{2}$ for $F$ in $H$ relative to $D$ such that
$T_{1} \preceq T_{2} \prec T_{3}$, then
$H(F,T_{1})=H(F,T_{3})$.

\item If $T_{1}$ is an initiation-point for $F$ in $H$
relative to $D$, and there is no termination-point
$T_{2}$ for $F$ in $H$ relative to $D$ such that
$T_{1} \prec T_{2} \prec T_{3}$, then
$H(F,T_{3})=true$.

\item  If $T_{1}$ is a termination-point for $F$ in
$H$ relative to $D$, and there is no initiation-point
$T_{2}$ for $F$ in $H$ relative to $D$ such that
$T_{1} \prec T_{2} \prec T_{3}$, then
$H(F,T_{3})=false$.

\item For all t-propositions  $F$ {\tt holds-at} $T$ in $D$, 
$H(F,T)=true$, 
and for all t-propositions 
``$\neg F$ {\tt holds-at} $T'$'' in $D$, $H(F,T')$ $=false$.
\end{enumerate}

\item A domain description $D$ is {\em consistent} iff it has a model.
Also, $D$ {\em entails} (written $\models$) the t-proposition 
$F \mbox{ {\tt holds-at} } T$
($\neg F \mbox{ {\tt holds-at} } T$, resp.), 
iff for every model $H$ of $D$, $H(F,T)=true$
($H(F,T)=false$, resp.). 
\end{itemize}
Note that the t-propositions, in effect,
are like ``static'' constraints that interpretations must satisfy
in order to be deemed models. 
We can extend the language ${\cal E}$ with {\em ramification} statements, called
{\em r-propositions}, of the form
$L \mbox{ {\tt whenever} } C$, where $L$ is a fluent literal
and  $C$ is a set of fluent literals. 
These are also understood as constraints on the interpretations,
but with the difference of being ``universal'', i.e.
applying to every time point.
Formally, the
definition of a model is extended with:

\begin{itemize}
\item[]
\begin{enumerate}
\item[5.]
For all r-propositions
$L \mbox{ {\tt whenever} } C$,  in $D$,
and for all time-points $T$,
if $H$ satisfies $C$ at $T$ then 
$H$ satisfies $\{L\}$ at $T$. 
\end{enumerate}
\end{itemize}
In addition, the complete formalization of ramification statements 
requires a suitable extension of the definitions of initiation- and
termination-point. The interested reader is refered to \cite{KaMi97bis} for 
the details. 

As an example,
consider the following  simple ``car engine domain''  $D_c$,
with action constants $\TurnOn$ and $\Empty$
and fluents $\Running$ and $\Petrol$:
\\

$\TurnOn \mbox{ {\tt initiates} } \Running \mbox{ {\tt when} }
\{\Petrol\}$ \hfill ($D_{c}$1)

$\Empty \mbox{ {\tt terminates} } \Petrol$ \hfill ($D_{c}$2)

$\TurnOn \mbox{ {\tt happens-at} } 5$ \hfill ($D_{c}$3)

$\Petrol \mbox{ {\tt holds-at} } 1$ \hfill ($D_{c}$4)
\\ 

\noindent
It is easy to see, for example, 
that $D_{c}$ entails $ \Running$ {\tt holds-at} $7$ and that 
$D_{c}$ extended via the h-proposition
$\Empty \mbox{ {\tt happens-at} } 3$ does not.


\section{ Planning with ${\cal E}$}
\label{planning-with-E}

The language ${\cal E}$ with its explicit reference to actions as
h-propositions in its basic ontology is naturally suited for the 
problem of planning.
Let a {\bf goal} be a set of t-propositions.
Then, given a domain description $D$ and a goal $G$, 
planning amounts to constructing a set $\Delta$ of h-propositions such that 
$D \cup \Delta$ entails  $G$.

In general, however, the extension of $D$ via 
the plan $\Delta$ might be required to
respect some given {\bf preconditions} for the actions in  $\Delta$.
These preconditions can be represented 
by a new kind of statements, called 
{\bf p-propositions} (``p'' for ``preconditions''), of the form
$A$ {\tt needs} $C$, 
where $A$ is
an action constant and $C$ is a 
non-empty 
set of fluent literals.
Intuitively, 
the fluents in $C$ are conditions that must hold at any time that 
the action $A$ is performed.
Note that, alternatively,  
preconditions could be encoded via additional conditions in 
{c-propositions} already appearing in the 
domain descriptions. 
The use of p-propositions is though simpler and more modular.

\begin{definition}
An {\bf ($\cal E$-)planning domain} is a pair $\langle D,P \rangle$,
where  $D$ is a domain description 
and $P$ is a set of p-propositions. 
\end{definition}
The semantic interpretation of the new type of sentences is 
that of integrity constraints on the domain descriptions.

\begin{definition}
Given a  planning domain $\langle D,P\rangle$, $D$ {\bf satisfies} $P$,
written  $ D \models P$,  iff for all
p-propositions $A$ {\tt needs} $C$ in $P$, and for all h-propositions 
$A$ {\bf happens-at} $T$ in $D$, $D$ entails $C(T)$, 
where $C(T)$ denotes the set of t-propositions obtained by transforming 
every fluent literal in $C$ into the respective t-proposition at 
$T$.
\end{definition}
The planning problem is then defined as follows.
\begin{definition}
Given a planning domain $\langle D,P \rangle$ and a goal 
$G$, a {\bf (safe) plan for} $G$ {\bf in} $D$ is a set $\Delta$ of h-propositions
such that $D \cup \Delta$ is consistent and :

$\bullet$  $D \cup \Delta \models G$,

$\bullet$  $D \cup \Delta \models P$.
\end{definition}
Note that the initial state of the planning problem is assumed to be
contained in the given domain description, and 
might amount to a set of t-propositions at some initial time point, or,
more generally 
a set of  t-propositions over several time points, not necessarily all
coinciding with a unique initial time point.

The above definition of (safe) plan provides
the formal foundation of the $\cal E$-planner.
It is easy to see that, through the properties of the 
model-theoretic semantics of $\cal E$, 
a safe plan satisfies the requirements that 
(i) it achieves the given goal, and (ii) it is executable.

As an example, 
let us consider the simple ``car engine planning domain'' 
$\langle D_c', P_c\rangle $, with $ D_c'$
consisting of statements ($D_{c}$1),
($D_{c}$2) and ($D_{c}$4) from the previous section~\ref{E-language}
as well as:
\\

$\Fill \mbox{ {\tt initiates} } \Petrol$ \hfill ($D_{c}$5)

$\neg \Running \mbox{ {\tt holds-at} } 1$ \hfill ($D_{c}$6)
\\
\\ 
and $P_c$ consisitng of the p-proposition
\\

$\Fill \mbox{ {\tt needs} } \neg\Running$  \hfill  ($P_{c}$1)
\\ 
\\ 
Let the goal be  $G= \Running$ {\tt holds-at} $T_{f}$
for some  (final) time $T_{f}$.
Then, a plan for $G$ is given by the
set $\Delta_1= \{ \TurnOn \mbox{ {\tt happens-at} } T_{1}\}$ where 
$T_{1} \prec T_{f}$. 
This is
a {\em safe} plan in the sense that if we add $\Delta_1$
to $D_{c}'$, then both the goal $G$ and the p-proposition in $P_{c}$ are entailed
by the augmented domain.

Consider now the domain $D_c''$ obtained from 
$D_c'$ by removing ($D_{c}$4).
Note that then $D_c''$ has incomplete (initial) information 
about 
$\Petrol$. 
Then, the above plan $\Delta_1$
is no longer a safe plan for $G$ as there is no guarantee that the
car will have petrol at the time $T_1$ when the $\TurnOn$ action is assumed
 to take place. 
A safe plan is now given by $\Delta_{2} = \{ \TurnOn \mbox{ {\tt happens-at} } 
T_{1}, \Fill \mbox{ {\tt happens-at} } T_{2} \}$ with 
$T_{2} \prec T_{1} \prec T_f$.
In the context of $D_c''$,
the original plan $\Delta_{1}$ will be called a {\em weak plan}.
A weak plan is a set of h-propositions such that
the extension it forms of the given domain description might not entail the given goal,
but there is at least one model of the augmented domain description 
in which the goal holds true.
A weak plan depends upon a set of assumptions, in the form of t-propositions,
such that, if
these assumptions were true (or could be made true) then the weak plan
would be (or would become) a safe plan.
In the example above 
$\Delta_{1}$ is weak as it
depends on the set of assumptions
$A = \{ \Petrol \mbox{ {\tt holds-at} } T_{1} \}$.
$\Delta_{2}$ is obtained from $\Delta_{1}$ by adding
the additional action $\Fill \mbox{ {\tt happens-at} } T_{2}$,
ensuring that $A$
is entailed by $D_{c}'' \cup \Delta_{2}$.
\begin{definition}
Given a planning domain $\langle D, P \rangle$ and a goal 
$G$, a {\bf weak plan for} $G$ {\bf in} $D$ is a set $\Delta$  of h-propositions
s.t. there exists a model $M$ of $D \cup \Delta$ where:

$\bullet$ $M \models G$, and

$\bullet$ $M \models P$.
\\
$\Delta$ is {\bf conditional} or {\bf depends on the set of assumptions}
$A$ iff $A$ is a set of t-propositions such that 
$\Delta$ is not a safe plan for $G$ in $D$ but it is 
a safe plan for $G$ in $D \cup A$.
\end{definition}
Note that a safe plan is always a weak plan and that 
if a weak plan is not conditional on any assumptions then it is
necessarily a safe plan. 

Computing conditional weak plans will form the
basis for computing safe plans in the $\cal E$-planner 
that we will develop in section~\ref{E-planner}.
%
%
In general, if we have a weak plan for a given goal $G$,
conditional on a set of assumptions $A$,
then the 
original planning problem for $G$ is reduced to the subsidiary problem of generating a 
plan for $A$. 
In effect, 
this process allows to actively fill in by further actions 
the incompleteness in the domain description.
 
However, in some cases we may have
incomplete information on fluents that can not be affected by
any further actions or at a time point (e.g. an initial time
point) before which we cannot perform actions. 
%
In this paper we will concentrate on incompleteness of this kind, and 
we will study how to appropriately generate safe plans from weak plans
despite the luck of information. 

Of course, this may not always be possible, but there are 
many interesting cases, 
such as the following 
``vaccine'' domain 
$D_v$,
where a safe plan exists:
\\

$\InjectA \mbox{ {\tt initiates} } \Protected \mbox{ {\tt when} }
\{\TypeO\}$  \hfill  {($D_{v}$1)}

$\InjectB \mbox{ {\tt initiates} } \Protected \mbox{ {\tt when} }
\{\neg \TypeO\}$ \hfill  ($D_{v}$2)
\\
\\
Here, the fluent $\TypeO$ cannot be affected by any of action 
(we cannot change the blood type) and 
although its truth value is not known 
(we have incomplete information on this) 
we can still generate a safe plan for the 
goal, $G= \Protected$ {\tt holds-at} $T_{f}$,
by performing both actions $\InjectA$ and 
$\InjectB$ before time $T_{f}$. 

\section{An Argumentation Formulation of ${\cal E}$}
\label{argumentation}

Argumentation has recently proved to be a unifying mechanism
for most  existing non-monotonic formalisms  \cite{Bondarenko,Dung95}.
In \cite{lpnmr99}, we have adapted the  $LPwNF$ \cite{DiKa95}
argumentation framework to provide an equivalent reformulation of the original
Language ${\cal E}$ presented in section~\ref{E-language}
and to develop a proof theory for computing entailment of t-propositions 
in domain descriptions.
This will form the computational basis for our ${\cal E}$-Planner.
In this section, we give a brief review of the argumentation 
formulation for ${\cal E}$ concentrating on the methods and
results that would be needed for the ${\cal E}$-Planner.
%

Let a {\em monotonic logic} be a pair $({\cal L},\vdash)$
consisting of a formal language $\cal L$ 
(equipped with a negation operator $\neg$) 
and a monotonic derivability notion $\vdash$ between sentences
of the formal language.
Then, an abstract {\em argumentation program}, relative to $({\cal L},\vdash)$),
is a  quadruple $(B,{\cal A},{\cal A'},<)$ consisting
of 

\begin{itemize}
\item 
a {\em background theory} $B$, i.e. a (possibly empty) set of sentences in ${\cal L}$,
\item 
an {\em argumentation theory}  $\cal A$, i.e. a set of sentences in ${\cal L}$
(the {\em argument rules}),
\item 
an {\em argument base} ${\cal A'} \subseteq {\cal A}$,
and 
\item 
a {\em priority relation}, $<$  on the
ground instances of the argument rules, where
$\phi < \psi$ means that $\phi$ has lower priority than $\psi$.
\end{itemize}
Intuitively, any subset of the argument base can serve as a 
non-monotonic extension of the (monotonic) background theory, 
if this extension satisfies some requirements.
The sentences in the background theory can be seen as non-defeasible
argument rules which must belong to any extension.
One possible requirement that extensions of the background theory must satisfy
is that they are {\em admissible}, namely that they are:

\begin{itemize}
\item {\em non-self-attacking} and

\item
able to {\em counterattack} any (set of) argument rules {\em attacking} it.
\end{itemize}
Informally, a set of argument rules from $\cal A$ {\em attacks} another such 
set if the two sets
are in conflict, by deriving in the underlying logic
complimentary literals $\lambda$ and $\neg \lambda$,  respectively,
and the subset of the attacking set (minimally) responsible
for the derivation of $\lambda$ is {\em not overall lower in priority}
than the subset of the attacked set (minimally) responsible
for the derivation of $\neg \lambda$. A set of rules A is of lower priority
than another set B if it has a rule of lower priority than some rule in B
and does not contain any rule of higher priority than some rule in B.

Then any given sentence $\sigma$ of $\cal L$ is a 
{\em credulous} ({\em sceptical}, resp.) non-monotonic consequence of  an
argumentation program
iff $B \cup \Delta \vdash \sigma $ for {\em some} ({\em all}, resp.) 
maximally admissible extension(s) $\Delta$ of the program.

A domain description $D$ 
without t-propositions can be translated
into an argumentation program $(B(D), \ARE, \ARE', <_{\E})$, 
referred to as $\PED$,
such that there is a one-to-one correspondance between: 

i) 
models of $D$
and maximally admissible sets of arguments of $\PED$;

ii) 
entailment in $\cal E$ and sceptical non-monotonic
consequences  of $\PED$.
\\
These equivalence results
continue to hold when $D$
contains t-propositions or r-propositions 
by simply considering only the admissible
sets  that confirm the truth of all 
such propositions in $D$.

The basic elements of the translation 
of domain descriptions $D$ into argumentation programs $\PED$
are as follows.
All individual h- and c-proposition translations
as well as the relationships between time-points 
are included in the background theory $B(D)$, so that   
for all time-points
$T$, $T'$ and action constants $A$, 

\begin{itemize}
\item $B(D) \PR T \prec T'$ iff $T \prec T'$

\item $B(D) \PR \HappensAt(A,T)$ iff $A \mbox{ {\tt happens-at} } T$ is in $D$,

\item for each c-proposition 
\\
$A \mbox{ {\tt initiates} }$ $F$  $\mbox{ {\tt when} }$ $ \{L_{1}, \ldots , L_{n}\}$ 
in $D$ 
\\
(resp. $A \mbox{ {\tt terminates} } F$ $\mbox{{\tt when} }$ 
$ \{L_{1}, \ldots , L_{n}\}$),  
\\
$B(D)$ contains the rule
\\
$\Initiation(F,t) \! \leftarrow \! \HappensAt(A,t),\!  \Lambda(L_{1}),\!  \ldots ,
\! \Lambda(L_{n})$
\\
($\Termination(F,t) \! \! \leftarrow \! \! \HappensAt(A,t), \! 
\ldots,
\! \Lambda(L_{n})$ resp.),
where  $\Lambda(L_{i}) = (\neg) \HoldsAt(F_{i},t)$ if $L_{i} = (\neg) F_{i}$, 
for some fluent constant $F_{i}$.
\end{itemize}
As an example, consider the domain description $D_c$ in section~\ref{E-language}.
Then,
$B(D_c)$ contains the fact $HappensAt(TurnOn,5)$
and $B(D_c)$ contains the rules

$Initiation(Running, t) \leftarrow $

\hfill $ HappensAt(TurnOn,t), \! \! HoldsAt(Petrol,t)$

$Termination(Petrol,t)  \leftarrow  HappensAt(Empty,t).$
\\
The remaining components of $\PED$ are independent of the chosen
domain $D$:

\begin{itemize}
\item $\ARE$ consists of 
\end{itemize}
{\em Generation rules}: 
\\
$\HoldsAt(f,t_{2}) \!\leftarrow \!  \Initiation(f,t_{1}), t_{1} \! \prec \! t_{2}$ 
\hfill $(\PGe[f,t_2;t_1])$
\\
$\neg \HoldsAt(f,t_{2})\! \!\leftarrow\!\! \!
        \Termination(f,t_{1}), t_{1} \!\! \!\prec\! \!\! t_{2}$ \hfill  $(\NGe[f,\! t_2;\! t_1])$
\\
\\
{\em  Persistence rules}: 
\\
$\HoldsAt(f,t_{2}) \! \leftarrow\!
        \HoldsAt(f,t_{1}), t_{1} \! \prec \! t_{2}$ \hfill  $(\PP[f,t_2;t_1])$\\
$\neg \HoldsAt(f,t_{2}) \! \! \leftarrow\! \!
        \neg \HoldsAt(f,t_{1}), t_{1} \! \! \prec \! \! t_{2}$ \hfill  $(\NP[f,t_2;t_1])$
\\

\noindent
\begin{tabular*}{\axiomwidth}{l@{\hspace*{\itemindent}\hspace*{\itemindent}
\hspace*{\itemindent}}l@{\extracolsep{\fill}}r@{}}
{{\em Assumptions}: } & {$\HoldsAt(f,t)$} & {$(\PA[f,t])$}\\
{} & {$\neg \HoldsAt(f,t)$} & {$(\NA[f,t])$}\\
\end{tabular*}

\begin{itemize}
\item $\ARE'$ consists of all the generation rules and assumptions only.

\item $<_{\cal E}$ is such that the effects of later events
take priority over the effects of earlier ones. Thus
persistence rules have lower priority than ``conflicting''
and ``later'' generation rules, and 
``earlier'' generation rules  have lower priority than ``conflicting''
and ``later'' generation rules.
In addition, assumptions have lower priority than ``conflicting'' generation rules.
For example, given the vocabulary of $D_c$ in section~\ref{E-language},
$\PA[Running,5]  <_{\cal E} \NGe[Running,5;3]$
and 
$\NGe[Running,7;3] <_{\cal E} \PGe[Running,7;5]$.
\end{itemize}
Given this translation of the language ${\cal E}$ a proof theory 
can be developed by adapting 
the abstract,
argumentation-based 
computational framework in \cite{KaTo99} 
to the argumentation programs $\PED$.
The resulting proof theory is defined in terms of {\em derivations of trees},
whose nodes are sets of arguments in $\ARE$ attacking the
arguments in their parent nodes.
Let $S_0$ be a (non-self-attacking) set of arguments in $\ARE'$ 
such that $B(D) \cup S_0 \PR (\neg) H \! oldsAt(F,T)$,
for some literal ($\neg$) $F \mbox{ {\tt holds-at} } T$  that we want to prove to be entailed by $D$ 
($S_0$ can be easily built by backward reasoning). 
Then, two kinds of derivations are defined:

\begin{itemize}
\item[-]
{\em Successful derivations}, building, from 
a tree consisting only of the root $S_0$, 
a 
tree whose root $S$ is an admissible subset of 
$\ARE'$ such that $S \supseteq S_0$.

\item[-]
{\em Finately failed derivations}, 
guaranteeing the absence of any admissible set of arguments 
containing $ S_0$.
\end{itemize}
Then, the given literal is entailed by $D$ if there exists a successful
derivation 
with inital tree consisting only of the root $S_0$ and,
for every set $S_0'$ of argument rules in $\ARE'$ such that
$B(D) \cup S_0'$ derives (in $\vdash$ the complement of the  given literal,
every derivation for $S_0'$ is finitely failed. 

This method is extended in the obvious way to handle 
 conjunctions of literals rather than individual
literals by choosing $ S_0$ and $S_0'$ appropriately.
Also when a domain $D$ contains 
t-propositions we simply 
conjoin these to the literals  $ S_0$ and $S_0'$.
A similar extension of requiring that all the r-propositions are 
satisfied together with the goal at hand 
is applied for the domains containing such ramification statements.

The details of the derivations are not needed for the purposes of this paper.
Informally,
both kinds of derivation incrementally consider all attacks (sets of arguments in $\ARE$) against 
$S_0$ and, whenever the root does not itself 
counterattack one of its attacks, a new 
a new set of arguments in $\ARE'$ that can attack back this attack 
is generated and added to the  root.
Then, the process is repeated,
until every attack has been counterattacked successfully (successful derivation)
by the extended root 
or until some attack cannot be possibly counterattacked by any extension of the root
(finitely failed derivations)
During this process, the counterattacks are chosen in such a way that they do not attack the root.
For example, for the domain $D_c$ in section~\ref{E-language},
given $S_0=\{{PG[Running,7;5],PA[Petrol,5]}\}$,
monotonically deriving  $H \! oldsAt(Running,7)$, a 
successful derivation is constructed as follows:

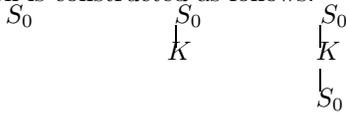
\begin{center}
\setlength{\unitlength}{0.00030000in}%
\begin{picture}(6424,1574)(4639,-1773)
\put(4050,-316){\makebox(0,0)[lb]{$\;S_0$}}
\put(6950,-316){\makebox(0,0)[lb]{$\;S_0$}}
\put(7100,-361){\line( 0,-1){375}}
\put(6950,-961){\makebox(0,0)[lb]{$K$}}
\put(9450,-316){\makebox(0,0)[lb]{$\;S_0$}}
\put(9576,-361){\line( 0,-1){375}}
\put(9500,-961){\makebox(0,0)[lb]{$K$}}
\put(9576,-1111){\line( 0,-1){375}}
\put(9500,-1750){\makebox(0,0)[lb]{$S_0$}}
\end{picture}
\end{center}

\noindent 
$S_0$ is attacked by $K= \{ NA[Petrol,5]\}$, trivially counterattacked
by $S_0$ itself.
Thus, in this simple example, no extension of the root is required.

\section{ The ${\cal E}$-Planner}
\label{E-planner}

The argumentation-based techniques discussed in the previous section
can be directly extended to compute plans for goals.
(In the sequel, we will sometimes mix the original
Language $\cal E$ formulation of problems and their corresponding formulation in the
argumentation reformulation.)
First, 
given a goal $G$,
in order to derive 
the (translation ($\Lambda(G)$ of the) goal in the underlying monotonic logic, 
a preliminary step needs to compute not only
a set of argument rules $S_0$,
but (possibly) also 
a set {\bf action facts} $\Delta_0 \subseteq {\cal H}$ where 

${\cal H}= \{ HappensAt(F,T) |$ $F$ is a fluent constant
and $T$ is a time-point$\}$ 
\\
$\Delta_0$ can be seen as a preliminary plan for the goal, that needs to be extended
first to a weak plan 
and then to a safe plan.
Every time a new action fact is added to a plan, 
any preconditions of the action need to be checked and, possibly, enforced,
by adding further action facts.
The computation of safe plans from weak ones
requires blocking, if needed, any 
(weak) plan for the complement of any literal in the goal.

The following is a high-level definition of
the  ${\cal E}$-Planner in terms of  the argumentation-based 
re-formulation of the $\cal E$-language:

\begin{definition}
\label{planner-def}
Given a planning domain $\langle D,P \rangle$
and a goal $G$, an {\bf $\cal E$-plan for} $G$ is a set 
h-propositions $\Delta^{\cal E}$  such
that  $\Delta^{\cal E}=\{ A \mbox{ {\tt happens-at} } T | HappensAt(A,T) \in \Delta\}$,
where 
$\Delta \subseteq {\cal H}$ 
is derived 
as follows:
\begin{itemize}
\item[1)] Find  a set of arguments $S_0 \subseteq \ARE'$ 
and a set of action facts $\Delta_{0} \subseteq {\cal H}$ 
such that $B(D)\cup\Delta_0\cup S_0 \vdash G$;
\item[2)] Construct a set of arguments 
$S \subseteq \ARE'$ and a set of action facts
$\Delta_{w} \subseteq {\cal H}$ such that
(i) $S_0 \subseteq S$
and $\Delta_{0} \subseteq \Delta_{w}$,
and 
(ii)
$S$ is an admissible set of arguments wrt the 
augmented argumentation program  $\NPED=(B(D')\cup\Delta_w,\ARE,\ARE',<_{\E})$, where 

$D'=D \cup\{\Lambda(C(T)) | A \mbox{ {\tt needs} } C \in P, HappensAt(A,T) \in \Delta_w\}$.

\item[3)] 
If 
every assumption in $S$ 
is a sceptical non-monotonic consequence of 
the augmented argumentation program $\NPED$ 
then 
$\Delta = \Delta_w$.
\item[4)] Otherwise, $\Delta$ is a set of action facts such that $\Delta_w \subset \Delta$
and:
\begin{itemize} 
\item[4.1)] For every set of arguments $R\subseteq \ARE'$ 
such that $B(D)\cup\Delta \cup R \vdash \neg G$,
where $\neg G$ stands for the complement of some literal in $G$,
there exists no $R' \subseteq \ARE'$ such that $R \subseteq R'$ and $R'$ is admissible
wrt the augmented argumentation program
$\NNPED=(B(D'')\cup\Delta,\ARE,\ARE',<_{\E})$, where

$D''=D \cup\{\Lambda(C(T)) | A \mbox{ {\tt needs} } C \in P, HappensAt(A,T) \in \Delta\}$.

\item[4.2)]  There exists a set $S'\subseteq \ARE'$ such that $S \subseteq S'$ and
$S'$ is admissible
wrt $\NNPED$.
\end{itemize} 
\end{itemize}
\end{definition}
In the first two steps, the ${\cal E}$-Planner computes a weak plan 
for the given goal. 
If this does not depend on any assumptions (step 3) then it is a safe plan,
as no plan for the complement of the goal is possible.
Otherwise (step 4),  the planner attempts to extend the weak plan 
in order to block the derivation of the complement,  $\neg G$, of the goal.
In order to do so,
it considers each possible set of arguments, $R$, which 
would derive $\neg G$ (in the augmented background theory)
and extends the plan so
that $R$ can not belong to any admissible set of the resulting theory.
A successful completion of step 4 means that the weak plan $\Delta_{w}$ 
has been rendered into a safe plan $\Delta$. 

The correctness of the planner is a direct consequence of the
correctness of the argumentation proof theory on which it is based.
\begin{theorem}
Given a planning domain $\langle D,P\rangle$ and a goal $G$,
let  $\Delta_w$ be the set of action facts
computed at step 2 in definition~\ref{planner-def}.
Then, the set 

$\Delta_w^{\cal E}$=$\{ A \mbox{ {\tt happens-at} } T | Happens(A,T) \in \Delta_w\}$
\\
is a weak plan for $G$.
\end{theorem}
{\bf Proof}: 
As every admissible set of arguments is contained in some maximally admissible set
\cite{lpnmr99} and, as discussed in section~\ref{argumentation},
every maximally admissible extension of the argumentation-based reformulation of a 
domain description in $\cal E$ corresponds to a model of the original domain,
$S$ computed at step 2 in definition~\ref{planner-def} 
corresponds to a model $M$ of $D'\cup\Delta_w^{\cal E}$ entailing $G$.
Because of the way t-propositions in domains are handled,
as additional conjuncts in goals,
this implies that  $M$ satisfies all preconditions of actions in $\Delta_w^{\cal E}$.
Thus, $M$ is a model of $D\cup\Delta_w^{\cal E}$ such that $M \models G$ and $M \models P$
and the theorem is proven.

The following theorem can be proven in a similar way:

\begin{theorem}
Given a planning domain $\langle D,P\rangle$ and a goal $G$,
let  $\Delta^{\cal E}$ be an $\cal E$-plan  for $G$.
Then, $\Delta^{\cal E}$ is a safe plan for $G$.
\end{theorem}
%
The high-level definition of the $\cal E$-Planner given above in definition~\ref{planner-def}
can be mapped onto a more concrete planner by suitably extending the 
argumentation-based proof theory proposed in \cite{lpnmr99}.
$\Delta_0$ can be computed directly 
while computing $S_0$, by an {\bf abductive process} which 
reasons backwards with the sentences in the background theory.
Also, 
one needs to define suitable:
\begin{itemize}
\item[-]
{\bf extended successful derivations},
for computing incrementally $\Delta_w$ from $\Delta_0$ at step 2
and the final $\Delta$ at step 4.2 from the extension of $\Delta_w$ computed at step 4.1;
\item[-]
{\bf extended finitely failed derivations},
for computing incrementally the required extension of $\Delta_w$ at step 4.1.
\end{itemize}
As the original derivations,
the extended ones incrementally consider all attacks against the root of the trees
they build and augment the root so that all such attacks are counterattacked,
until every attack has been counterattacked (successful derivations)
or some attack cannot be counterattacked (failed derivations). 
Again, all nodes of trees are sets of arguments.

In addition, 
both new kinds of derivation are integrated with abduction
to generate action facts so that success and failure are guaranteed, respectively.
The action facts are chosen to allow for counterattacks to exist (successful derivations)
or for additional attacks to be generated (failed derivations).
Thus, extended successful derivations return both a set of argument rules in $\ARE'$ 
and a set of action facts,
wherever extended failed derivations just return a set of action facts
(the ones needed to guarantee failure).

Both kinds of derivations need to add to the given background theory (domain)
the t-propositions that are preconditions of any abduced action.
By the way t-propositions are handled, this amounts to extending dynamically the given goal
to prove or disprove, respectively.

Finally, both kinds of derivation require the use of {\bf suspended nodes},
namely that could potentially
attack or counterattack their parent node
if some action facts were part of the domain.
These nodes become actual attacks and counterattacks if and when the action facts 
are added to the accumulated set.
If, at the end of the derivations,
these action facts are not added,
then suspended nodes remain  so and do not affect the outcome of the derivations.

Let us illustrate the intended behaviour of the extended derivations with a simple example.
Consider the simple ``car engine'' domain $D_c'$ in section~\ref{planning-with-E}, and 
let 
$G= HoldsAt(Running,T_{f})$ for some fixed final time $T_{f}$.
We will show how the safe $\Delta_w=\{HappensAt(TurnOn,T_{1})\}$,
with $ T_{1} \prec T_{f} $,  is computed.

\begin{itemize}
\item[1)] $S_{0}= \{PG[Running,T_{f};T_{1}], PA[Petrol,T_{1}]\}$ 
and 
\\
$\Delta_{w} = HappensAt(TurnOn,T_{1})$, with $ T_{1} \prec T_{f}$.

\item[2)] The only possible attack against $S_{0}$ is $\{ NA[Petrol,T_{1}]\}$,
which  is trivially counterattacked  by $S_{0}$ itself. Thus $S_{0}$ is admissible.

\item[3)] Let us examine the only assumption $PA[Petrol,T_{1}]$ in $S_{0}$,
and try to prove that it holds in all admissible extensions
of the given domain extended by $\Delta_{w}$.
Consider the complement $NA[Petrol,T_{1}]$ of the assumption, and 
let us prove that it holds in no admissible extension.
This can be achieved by an ordinary finitely failed derivation
(without abducing any additional action fact in order to do so),
as there is an attack, $\{ PP[Petrol,T_{1};1], PA[Petrol,1]\}$,
against the above complement, 
which cannot be counterattacked. 
\end{itemize}
Thus, $\Delta_w$  is a safe plan for $G$.
Consider now the domain $D_c''=D_c'-\{(D_c4)\}$. Then, step 3 above fails 
to prove that the given assumption is a sceptical non-monotonic consequence of the 
augmented domain, and thus $\Delta_w$ is just a weak plan.

\begin{itemize}
\item[4.1)]  $\neg G= \neg HoldsAt(Running,T_{f})$ is derivable 
via 
$R=\{NA[Running,T_{f}] \}$.
This is attacked by $\{PG[Running,T_{f};T_{1}]$, $PA[Petrol,T_{1}]\}$,
which is counterattacked by $\{NA[Petrol,T_{1}]\}$, which, if added to $R$,
would provide an admissible extension in which $\neg G$ holds. 
An extended finitely failed derivation can be constructed to prevent this as follows:
the extended root $R\cup\{NA[Petrol,T_{1}]\}$ is attacked by 
$\{ PG[Petrol,T_{1};T_{2}] \}$ if $\Delta_w$ is augmented to give 
$\Delta=\Delta_w \cup \{ HappensAt(Fill,T_{2}) \}$, with $ T_{2} \prec T_{1}$.
As the new attack cannot be counterattacked, the derivation fails.
\end{itemize}
Thus,  $\Delta$ is a safe plan for $G$ (for simplicity we omit step 4.2 here).

Note that the method outlined above relies upon 
the explicit treatment of non-ground arithmetical constraints
over time-points (see \cite{aclpnmr98}). 

\section{Incomplete planning problems}

In this section we will illustrate through a series of examples the
ability of the ${\cal E}-Planner$ to produce safe plans
for incompletely specified problems. 
In particular, we will consider problems where the
incompleteness on some of the fluents is such that it 
cannot be affected by actions and hence 
our knowledge of them cannot be
(suitably) completed by adding action facts to plans.

Let us consider again the ``vaccine'' domain $D_v$ 
at the end of section~\ref{planning-with-E},
and the goal  $G= \{ \Protected$ {\tt holds-at} $T_{f}\}$,
for some final time $T_{f}$.
We will show how, in this example, the $\cal E$-planner
reasons correctly with
the ``excluded middle rule''  to produce a safe plan for $G$,
despite the fact
that it is not known whether the blood if of $Type0$ or not.
A weak plan for the goal is given by 
$\Delta_w = \{ HappensAt(InjectA, T_{1}) \}$ with $ T_{1} \prec T_{f}$. 
Indeed: 
given $S_0=\{ PG[Protected,T_{f};T_{1}], PA[Type0,T_{1}] \}$,
$B(D_{v}) \cup S_0 \cup \Delta_w \vdash G$, 
$S_0$ is admissible (steps 1 and 2).
The plan $\Delta_w $ is conditional 
on the set of assumptions $ \{ PA[Type0,T_{1}] \}$ (step 3).
Note that there is no action that can affect the fluent $Type0$,
so $\Delta_w$ cannot be extended so that it can derive the assumption.
Nevertheless, we can extend $\Delta_w$ to a safe plan $\Delta$
by constructing failed derivations for 
$\neg G=\{\neg HoldsAt(\Protected,T_{f})\}$, 
as illustrated below (step 4.1).

The only way to derive $\neg G$ is by means of
the set of arguments $R_1=\{ NA[Protected, T_{f}] \}$.
This is attacked by $S_0$ itself, which 
can be counterattacked (only) if
the root $R_1$ is extended via the assumption
$ NA[Type0,T_{1}]$.

The initial root $R_1$ and thus
the extended root $R_2$ are attacked by the set of arguments
$\{PG[Protected,T_{f};T_{2}], NA[Type0,T_{2}]\}$
with $ T_{2} \prec T_{f}$,
provided we add to $\Delta_w$ the action fact
$HappensAt(InjectB, T_{2})$ to give a new plan 
$\Delta = \{ HappensAt(InjectA, T_{1}), HappensAt(InjectB, T_{2})\}$.
In order to successfully counterattack the new attack we need to add
further to the root the argument $PA[Type0,T_{2}]$,
obtaining a new root $R_3$.

$R_3$ is (newly) attacked by 
$\{ PP[Type0,T_{1};T_{2}]$, $PA[Type0,T_{2}] \}$, if
$ T_{2} \prec T_{1}$,
and by 
$\{ NP[Type0,T_{2};T_{1}]$, $NA[Type0,T_{1}] \}$, if
$ T_{1} \prec T_{2}$.
(Note also that necessarily 
$ T_{1} \neq T_{2}$ as otherwise $R_3$ would attack itself.)
These attacks can only be counterattacked 
via
one generation rule
for $\neg HoldsAt(Type0,T_{1})$ and one for 
$HoldsAt(Type0,T_{2})$, respectively.  
But
no such generation rules are possible.

This concludes the construction of the 
only required finitely failed extended derivation for $\neg G$.
(Again, we omit step 4.2.)
The computed $\cal E$-plan  $\Delta$ 
is indeed a safe plan for $G$. 
Note that no p-propositions are present in this example
and thus no extended domain is generated.
%

The following example illustrate the use of
observations (in the form of t-propositions)
to provide some (partial) implicit information on the domain.
Consider the domain $D_{i}$: 
\\

$\InjectC \mbox{ {\tt initiates} } \Protected \mbox{ {\tt when} }
\{\TypeA\}$ \hfill  ($D_{i}$1)

$\InjectD \mbox{ {\tt initiates} } \Protected \mbox{ {\tt when} }
\{\Weak\}$ \hfill ($D_{i}$2)

$\Bite \mbox{ {\tt initiates} } \Infected \mbox{ {\tt when} }
\{\TypeA\}$\hfill ($D_{i}$3)

$\Expose \mbox{ {\tt initiates} } \Infected \mbox{ {\tt when} }
\{\Weak\}$ \hfill ($D_{i}$4)

$\neg \Infected \mbox{ {\tt holds-at} } 1$ \hfill ($D_{i}$5)

$\Infected \mbox{ {\tt holds-at} } 4$ \hfill ($D_{i}$6)
\\
\\
and the goal  $G= \Protected$ {\tt holds-at} $T_{f}$, for some 
final time $4 \prec T_{f}$.
The fluents $\TypeA$ and $\Weak$ are incompletely specified.
but the observations (t-propositions)
essentially give indirectly the information that either 
$\TypeA$ or $\Weak$ must hold. This  then allows, similarly to the previous example, 
to generate a safe plan for $G$ by applying
both actions $\InjectC$ and $\InjectD$.

A weak plan $\Delta_w = \{ HappensAt(InjectC, T_{1}) \}$ with $ T_{1} \prec T_{f}$
is first generated (steps 1 and 2), conditional on the assumption set
$\{ PA[TypeA,T_{1}] \}$ (step 3). Then, step 4.1 generates
$R = \{ NA[Protected, T_{f}] \}$, proving $\neg G$.
$R$ is attacked by
$\{ PG[Protected,T_{f};T_{1}], PA[TypeA,T_{1}] \}$ and can only 
be defended if $R$ is extended to 
$R' = R \cup \{ NA[TypeA,T_{1}] \}$, which is admissible. 
However, $R'$ needs to be extended to confirm the t-propositions.
To confirm $(D_i5)$,
the assumption $NA[Infected,1]$ needs to be added to $R'$.
Moreover, to confirm $(D_i6)$,
we need to add either $R_1=\{ PG[Infected,T_{f};T_{2}], PA[TypeA,T_{2}] \}$ or 
$R_2=\{ PG[Infected,T_{f};T_{2}], PA[Weak,T_{2}] \}$,
with $T_{2} \prec 4$.
But adding the fist such set would render the resulting set 
non-admissible (as both
$PA[TypeA,T_{2}]$ and $NA[TypeA,T_{1}]$ belong to it).
Hence, the only viable extension is $R''=R'\cup R_2 \cup \{NA[Infected,1]\}$.
This set is admissible (if we also abduce 
the action  $HappensAt(Expose, T_{2})$).
To prevent that, 
we find an attack that cannot be counterattacked successfully:  
$ \{ PG[Protected,T_{f};T_{3}], PA[Weak,T_{3}] \}$, $T_{3} \prec T_{f}$,
extending $\Delta_{w}$ to 
$\Delta = \Delta_{w} \cup \{ HappensAt(InjectD, T_{3}) \}$.
This attack can only be counterattacked by adding the assumption $NA[Weak,T_{3}]$,
rendering the root self-attacking (as $ PA[Weak,T_{2}]$ belongs to it).
Thus, $\Delta$ is a safe plan.

The next example shows how the ${\cal E}$-planner exploits 
ramification information to generate safe plans for incompletely
specified problems. Let 
$D_{r}$ be:
\\

$\InjectB \mbox{ {\tt initiates} } \Protected \mbox{ {\tt when} }
\{\neg \TypeO\}$  \hfill ($D_{r}$1)

$\InjectE \mbox{ {\tt initiates} } \Protected \mbox{ {\tt when} }
\{\Strong\}$ \hfill ($D_{r}$2)

$\Strong \mbox{ {\tt whenever} } \{\TypeO\}$  \hfill ($D_{r}$3)
\\
\\
and 
$G= \Protected$ {\tt holds-at} $T_{f}$, for some 
final time $4 \prec T_{f}$. 
The fluents $\TypeO$ and $\Strong$ are incompletely specified. 
The ramification statement requires that 
either $\neg \TypeO$ or $\Strong$ must hold (at any 
time).
Then, similarly to the above examples, the ${\cal E}-planner$ can generate
the safe plan 
$\{ HappensAt(InjectB, T_{1})$, 
$HappensAt(InjectE, T_{2})\}$, with $ T_{1}, T_{2} \prec T_{f}$.
During the computation of this plan,
to render the only proof
of $\neg G$ non-admissible we generate, in addition to the attack
given by the weak plan $\{HappensAt(InjectB, T_{1})\}$, an extra
attack by adding the action $HappensAt(InjectE, T_{2})$.
The first attack can only be counterattacked by $\{PA[Type0,T_{1}]\}$ 
and the second only by $\{NA[Strong,T_{2}]\}$. As 
these assumptions persist, it is not possible to satisfy the
ramification statement at any time between $T_{1}$ and $T_{2}$.
Hence there is no admissible extensions that can prove $\neg G$
and the plan is safe.

\section{Conclusions}

We have shown how we can formulate planning within the framework of 
the Language ${\cal E}$  and have used the argumentation reformulation 
of this framework to define a planner that is able to solve 
problems with incomplete information. 

A planner with similar aims has been defined in \cite{reiter99}.
Both this planner and our $\cal E$-planner 
regress to a set of assumptions
which, when entailed by the incomplete theory, guarantees the plan to be safe.
However, \cite{reiter99} uses a classical theorem prover 
to check explicitly this entailment at the initial situation
(in general, the required entailment is the non-monotonic entailment of the action framework
in which the planning problems are formulated). 
Instead, in the $\cal E$-planner
the incompleteness, and hence the assumptions to which one regresses, 
need not refer to the initial state only.
Moreover, the ${\cal E}$-planner uses these assumptions in the 
computation of the initial possibly weak plan and then to help in the 
extension of this to a safe plan. We are studying 
other planning algorithms (within the same argumentation
formulation of the language ${\cal E}$) which use more actively
the assumptions on which weak plans are conditional. One such 
possibility is to  try to extend
the plan so as to re-prove the goal but now assuming a-priori the 
contrary of these assumptions.
The search space of this type of planning 
algorithm is different and comparative studies of effeciency can be made. 

\cite{smithweld98}
introduce the notion of {\em conformant planning}
for problems with incomplete information about the
initial state and for problems where the outcome of actions may be uncertain.
Our safe plans correspond to conformant plans for problems of the first type. 
The emphasis of this work is on the 
development of an efficient extension of Graphplan to 
compute conformant plans,  by first considering all possible worlds
and, in each world, all possible plans, and then extracting a 
conformant plan by considering the interactions
between these different plans and worlds. 

\cite{giunchigliaE-99}
considers the problem of planning with incomplete information
on the initial state within the action language ${\cal C}$.
This is an expessive language that allows concurrent and 
non-deterministic actions together with ramification and
qualification statements.
Our safe plans correspond to the notion of {\em valid plans}
which in turn are conformant plans. 
To find a valid plan, the ${\cal C}$-Planner generates a possible 
plan and then tests, using a SAT solver,
whether the generated plan can be executed in 
all the possible models.
Possible plans can be seen as weak plans, but 
we allow in the ``testing phase'' for the dynamic expansion 
of the weak plan into a safe plan. 

A general difference 
with both \cite{smithweld98,giunchigliaE-99} is that 
the ${\cal E}$-planner is goal-oriented, with an active 
search for actions both for the satisfaction of the goal
and for ensuring that the generated plan is executable in any
of the many possible worlds for the problem.

Another difference is at the level of expressivenes, in that 
we allow observations (not only at an initial state)
and can exploit indirect information given by them to help handle the
incompleteness.

We are currently developing an implementation of the ${\cal E}$-planner 
based on an earlier implmentation of the language ${\cal E}$ and aim to
carry out experiments with standard planning domains.
In this initial phase of our study
we have not considered efficiency issues, concentrating
specifically on the problem of planning under incompletness. In future 
work we need to address these issues by studying the problem of 
effective search in the space of solutions. 
One way to do this is to consider the integration of constraint 
solving in the planner as in constraint logic programming 
and its extension with abduction \cite{aclpnmr98,slp98}.

We are considering several extensions of the ${\cal E}$-planner to allow for more general plans
e.g. containing non-deterministic actions (or actions with uncertain effects).
These extensions  would require corresponding 
extensions of the expressiveness of the Language ${\cal E}$. 
Also, the extension of the ${\cal E}$-planner to accommodate sensing,
in the form of accepting further observations (t-propositions) in the 
problem description, is a natural problem for future work.

\section*{Acknowledgements}
This research has been partially supported by the 
EC Keep-In-Touch Project
``Computational Logic for Flexible Solutions to Applications''.
The third author has been supported by the
UK EPSRC Project ``Logic-based multi-agent systems''.

\bibliography{E-Planner}
\bibliographystyle{aaai}

\end{document}